\documentclass[conference]{IEEEtran}
\IEEEoverridecommandlockouts
\usepackage{cite}
\usepackage{amsmath,amssymb,amsfonts}
\usepackage{tabularx}
\usepackage{algorithm, algorithmicx}
\usepackage[noend]{algpseudocode}
\usepackage{graphicx}
\usepackage[normalem]{ulem} 
\usepackage{textcomp}
\usepackage{xcolor}
\usepackage{xspace}
\usepackage{booktabs}
\usepackage{subcaption}
\usepackage{multirow}
\usepackage{listings}
\usepackage{enumitem}
\usepackage{float}
\usepackage{tcolorbox}
\usepackage{color}
\usepackage{framed}
\usepackage{url}

\newcommand{\Approach}[1]{\texttt{RT-LM}}

\def\BibTeX{{\rm B\kern-.05em{\sc i\kern-.025em b}\kern-.08em
    T\kern-.1667em\lower.7ex\hbox{E}\kern-.125emX}}

\definecolor{shadecolor}{rgb}{0.92,0.92,0.92}

\begin{document}

\title{\texttt{RT-LM}: Uncertainty-Aware Resource Management for Real-Time Inference of Language Models}

\author{
Yufei Li$^{1}$
\xspace\xspace\xspace    
Zexin Li$^{1}$
\xspace\xspace\xspace   
Wei Yang$^{2}$
\xspace\xspace\xspace  
Cong Liu$^{1}$
\\
$^{1}$University of California, Riverside
\xspace\xspace\xspace 
$^{2}$University of Texas at Dallas
\\ 
$^{1}$\texttt{\{yli927,zli536,congl\}@ucr.edu}
\xspace\xspace\xspace
$^{2}$\texttt{wei.yang@utdallas.edu}
}

\maketitle

\begin{abstract}
Recent advancements in language models~(LMs) have gained substantial attentions on their capability to generate human-like responses.
Though exhibiting a promising future for various applications such as conversation AI, these LMs face deployment challenges on various devices due to their extreme computational cost and unpredictable inference latency.
Such varied inference latency, identified as a consequence of uncertainty intrinsic to the nature of language, can lead to computational inefficiency and degrade the overall performance of LMs, especially under high-traffic workloads.
Unfortunately, the bandwidth of these uncertainty sources is extensive, complicating the prediction of latency and the effects emanating from such uncertainties.
To understand and mitigate the impact of uncertainty on real-time response-demanding systems, we take the first step to comprehend, quantify and optimize these uncertainty-induced latency performance variations in LMs.
Specifically, we present \Approach{}, an uncertainty-aware resource management ecosystem for real-time inference of LMs. \Approach{} innovatively quantifies how specific input uncertainties, recognized within the NLP community, adversely affect latency, often leading to an increased output length. Exploiting these insights, we devise a lightweight yet effective method to dynamically correlate input text uncertainties with output length at runtime. Utilizing this quantification as a latency heuristic, we integrate the uncertainty information into a system-level scheduler which explores several uncertainty-induced optimization opportunities, including uncertainty-aware prioritization, dynamic consolidation, and strategic CPU offloading.
Quantitative experiments across five state-of-the-art LMs on two hardware platforms demonstrates that \Approach{} can significantly reduce the average response time and improve throughput while incurring a rather small runtime overhead.
\end{abstract}


\section{Introduction}

The recent surge in the development and dissemination of language models~(LMs) such as ChatGPT has significantly reshaped the landscape of natural language processing~(NLP)~\cite{brown2020language,chen-etal-2021-revisiting,chen-etal-2022-generate,li-etal-2022-share}.
This advancement holds immense promise for a multitude of applications, including multi-lingual robots and voice control devices integral to the future of smart homes~\cite{dialogue_hri,voice_gen,household_dialogue}.
Despite the impressive capability to generate human-like responses, these state-of-the-art LMs present a formidable challenge when attempting to deploy them on various devices due to their complex computational behaviors and unpredictable real-time inference capabilities~\cite{tay2022efficient, transformers}.
With the increasing demand for real-time language processing, server-backed systems, such as online chatbots (e.g., ChatGPT manages over 10 million daily queries) and live-translation services, exemplify the need for devices that can efficiently process simultaneous requests from multiple users, especially during peak times.

A set of recent works seek to enhance the inference latency of on-device LMs by crafting an array of model optimization techniques, including quantization~\cite{xu2022survey}, pruning~\cite{he2018multi,rethink_pruning}, and distillation~\cite{sanh2019distilbert}. 
These techniques aim at decreasing model complexity (thus the computational demand) while preserving their accuracy.
Nonetheless, a knowledge gap persists in understanding and exploring the correlation between an input text and the corresponding inference latency within a given LM from a system-level perspective.

The NLP community has recently brought to light various sources of uncertainties~\cite{local-coherence-discourse,LM_wsd,gunasekara-etal-2021-using-question,open-qa,zhu-etal-2020-question}, which have been shown to negatively impact model's accuracy and may introduce significant variations in the lengths of generated responses.
Take, for example, a broad and ambiguous question such as ``\textit{Can you tell me the history of art?}". 
This could prompt a LM to generate lengthier outputs, given that the history of art spans millennia and includes a multitude of cultures, styles, periods, and artistic movements.
Intuitively, the longer output a LM generates, the greater the inference latency, as each output token is sequentially generated with negligible computational difference~\cite{seq2seq-rnn,li-etal-2018-seq2seq}.
These sources of uncertainties, often intrinsic to the nature of language understanding and generation, can stem from varying data distributions~\cite{he-etal-2020-towards,kong-etal-2020-calibrated}, intricate model architectures~\cite{gal-16-dropout}, or even the non-deterministic parallel computing behaviors at runtime~\cite{niu2020real}, rendering the induced latency more complex and challenging to manage.
Consequently, it is critical to understand and mitigate such uncertainties due to their potential to induce non-trivial inference latency and computational inefficiency, or even hinder the prompt delivery of dialogue generation~(DG) due to degraded system performance.

This work is specifically motivated by the following queries: (i) What is the intrinsic correlation between an input text's uncertainty characteristics and the subsequent computational demand (and thus, the inference latency) for a given LM, such as why two syntactically similar inputs may necessitate dramatically different inference latencies? (ii) Is it feasible to devise a lightweight approach to predict an input's computational demand at runtime? and (iii) Can the system-level resource manager exploit these quantified input characteristics to improve latency performance during inference?
Understanding the quantifiable correlation between an input text and its computational demand is critical, as it could unveil novel opportunities for system-level optimization, thereby enhancing the performance and efficiency of LMs deployed on embedded devices, e.g., by deferring the execution of inputs with high computational demand thus reducing head-of-line blocking.

Our research attempts to comprehend, quantify, and optimize these uncertainty-induced variations on latency performance in LMs.
We propose a cohesive ecosystem that integrates an application-level uncertainty quantification framework with a system-level uncertainty-aware resource manager.
The application-level framework aims to precisely quantify task uncertainties and their potential impacts on latency.
Simultaneously, the system-level resource manager utilizes the provided estimations to make informed decisions on resource allocation and task scheduling, thereby mitigating the detrimental effects of uncertainties on system performance.

\noindent \textbf{Contributions.}
In this paper, we propose an uncertainty-ware resource management ecosystem, namely \Approach{}, for real-time on-device LMs.
Specifically, \Approach{} features three technical novelties:
1) It first quantitatively reveals how major input uncertainties---well-defined by the NLP community---negatively impact latency. Our findings demonstrate that uncertainty characteristics of an input text may notably increase the output length, i.e., the number of tokens in the generated response;
2) Building on this insight, we develop a lightweight yet effective method that can quickly correlate and quantify the output length for an input text at runtime, considering a comprehensive set of uncertainties defined by the NLP community;
3) Leveraging this quantification as a heuristic of latency, we incorporate the uncertainty information of each input into system-level scheduler that performs several optimizations, including uncertainty-aware prioritization, dynamic consolidation, and strategic utilization of CPU cores.

We implement \Approach{} mainly on an edge server.
We evaluate the response time and throughput across five state-of-the-art LMs\footnote{While there are larger models like ChatGPT that offer impressive capabilities, their resource-intensive nature makes them less viable for deployment.}, namely DialoGPT~\cite{dialogpt}, GODEL~\cite{GODEL}, BlenderBot~\cite{blenderbot}, BART~\cite{lewis-etal-2020-bart}, and T5~\cite{T5}.
We utilize \Approach{} four widely-researched benchmark datasets: 
\textit{Blended Skill Talk}~\cite{bst}, \textit{PersonaChat}~\cite{personachat}, \textit{ConvAI2}~\cite{convai2}, and \textit{Empathetic Dialogues}~\cite{empathetic_dialogue}.
For both the models and datasets, we use the versions released by Hugging Face.

Evaluation results demonstrate that \Approach{} achieves:
\begin{itemize}[leftmargin=*]
    \item \textbf{Efficiency}: \Approach{} outperforms all compared methods by a significant margin in most cases, improving the maximum response time by up to 30\% and throughput by up to 40\% compared to uncertainty-oblivious baselines.
     \item \textbf{Efficacy across a range of behaviors}: The tested workloads include five LMs with diverse task uncertainty characteristics and varied workload settings.
    \item \textbf{Robustness under malicious scenarios}: \Approach{} is resilient when facing adversarial conditions, effectively mitigating the impact of malicious tasks by resource management.
    \item \textbf{Runtime overhead}: The design and implementation of \Approach{} is efficient, incurring a rather small runtime latency and memory usage.
\end{itemize}

\section{Background and Challenges}

\subsection{Dialogue Generation using LMs}

Recently, pre-trained LMs such as ChatGPT and GPT-4~\cite{brown2020language} have emerged as a dominant force in the field of dialogue generation~(DG).
These models are characterized by their large size and are often trained on vast amounts of textual data, which demonstrate remarkable capabilities in understanding and generating human-like responses across a wide range of tasks.
A key property of these models is the \textit{autoregressive} generation process~\cite{transformers}, where output tokens are generated sequentially with each new token being conditioned on the previously generated tokens.
Consequently, the output length plays a pivotal role in determining the inference latency of a LM, as generating longer sequences inherently requires more time.
Depending on the nature of inputs, a LM may generate outputs of varied lengths.
For instance, a query that has clear and concise meanings may elicit a brief response, whereas an ambiguous or broad query may demand a considerably longer output.
This variability, often called \textit{linguistic uncertainty}~\cite{shelmanov-etal-2021-certain} by the NLP community, in output length and the subsequent impact on latency, can pose significant challenges when deploying LMs on resource-constrained devices, as the performance requirements and computational constraints must accommodate a wide range of potential latencies.

\subsection{Sources and Impacts of Linguistic Uncertainty}
\label{sec:uncertainty type}

\begin{table}[t]
\centering
\caption{Types of linguistic uncertainty, their definitions and example statements or questions.}
\footnotesize
\begin{tabularx}{0.5\textwidth}{>{\hsize=0.135\hsize}X>{\hsize=0.46\hsize}X>{\hsize=0.305\hsize}X}
\toprule
\textbf{Type} & \textbf{Definition} & \textbf{Statement/Question} \\
\midrule
Structural ambiguity & Uncertainty related to multiple possible parse structures, leading to outputs with varying lengths. & ``John saw a boy in the park with a telescope." \\
\midrule
Syntactic ambiguity & Uncertainty arising from multiple part-of-speech tags of a word, resulting in different interpretations. & ``Rice flies like sand." \\
\midrule
Semantic ambiguity & Uncertainty stemming from words with multiple meanings, leading to varying interpretations. & ``What's the best way to deal with bats?" \\
\midrule
Vague expressions & Uncertainty arising from broad concepts or highly-generalized topics that demand specific analysis. &      ``Tell me about the history of art.''  \\
\midrule
Open-endedness & Questions or statements that lack a single definitive answer and require providing relevant context, background, and explanations. & ``What are the causes and consequences of poverty in developing countries?" \\
\midrule
Multi-partness & Questions or statements containing multiple sub-questions or topics, which demand detailed answers. & ``How do cats and dogs differ in behavior, diet, and social interaction?" \\
\bottomrule
\end{tabularx}
\label{tab:linguistic_uncertainty}
\vspace{-0.3cm}
\end{table}

Linguistic uncertainty is a challenging and diverse sub-domain in NLP, which often leads to multiple interpretations of inputs and potentially varied outputs in dialogue systems.
The language and linguistics community has well-defined a categorization of linguistic uncertainty that encompasses the majority of uncertainty sources, including three types of lexical ambiguity (structural ambiguity~\cite{correfer-resolution,local-coherence-discourse}, syntactic ambiguity~\cite{e2e-corref-resolution,LM_wsd}, semantic ambiguity~\cite{ambiguous-cue-phrases,lebanoff-liu-2018-automatic}), vague expressions~\cite{gunasekara-etal-2021-using-question}, open-ended questions~\cite{mao-etal-2021-eliciting,open-qa}, and multi-part questions~\cite{zhu-etal-2020-question} that demand comprehensive answers and additional explanations.
Their definitions and example statements or questions are listed in Table~\ref{tab:linguistic_uncertainty}.

\section{Key Observations and Ideas}

\subsection{Uncertainty-Induced Negative Impact on LM Latency and the Root Cause}

\begin{figure}
    \centering
    \includegraphics[width=0.49\textwidth]{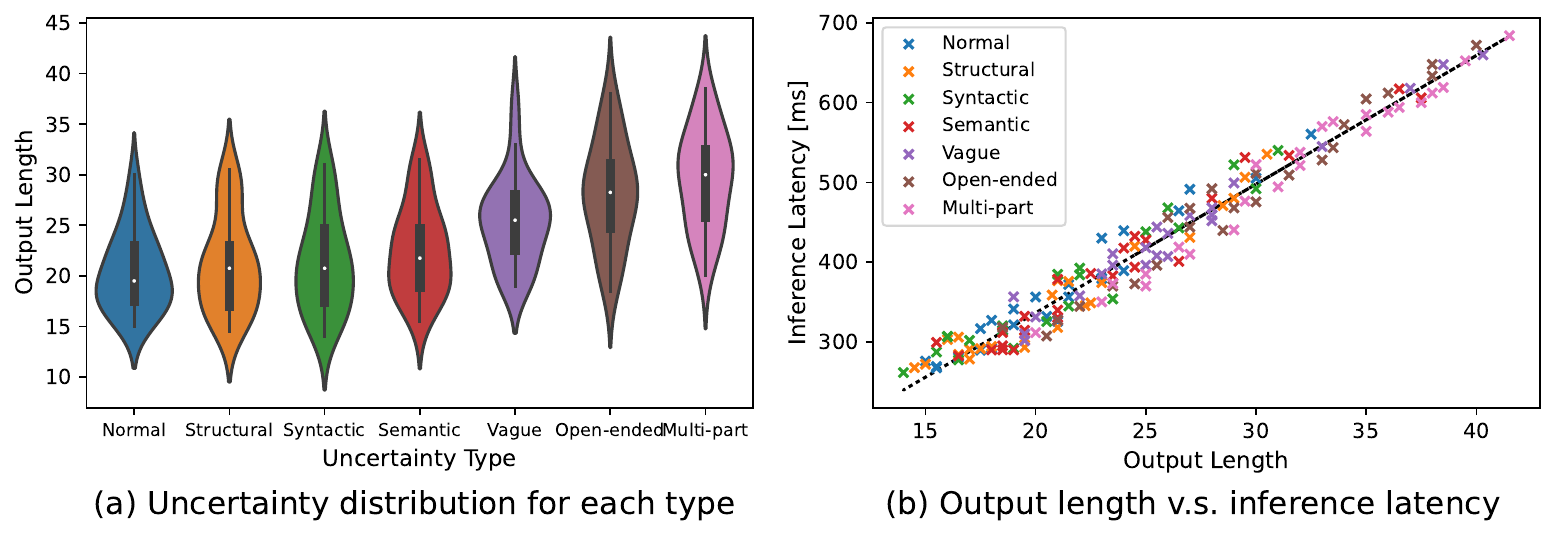}
    \caption{Observations of (a)~distribution of LM output lengths for inputs with different uncertainty types, and (b)~the correlation between LM output lengths and inference latency.}
    \label{fig:uncertainty_latency}
\end{figure}

\begin{figure*}
    \centering
    \includegraphics[width=0.98\textwidth]{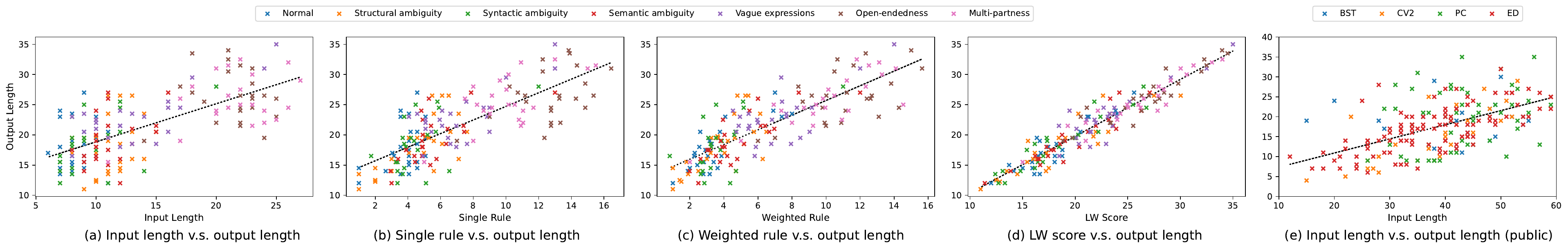}
    \caption{Correlation between average output length across the five LMs and (a)~input length, (b)~single rule-based score, (c)~weighted rule-based score, (d)~LW model scores for self-generated sentences that contain different types of uncertainties, as well as (e)~input length for sentences from the four benchmark datasets.}
    \label{fig:rule-length}
\end{figure*}

We conducted a comprehensive set of studies investigating the correlation between inputs' uncertainty characteristics and the resulting inference latency of several LMs.
Specifically, we create 1,000 utterances for each of the six uncertainty types (defined in Sec.~\ref{sec:uncertainty type}) and record the averaged output length as well as inference latency across DialoGPT, GODEL, BlenderBot, BART, and T5, as shown in Fig.~\ref{fig:uncertainty_latency}a. 
We observe that all types of linguistic uncertainties lead to longer outputs and non-trivially larger latencies to varying degrees.
Specifically, vague expressions, open-endedness, and multi-partness are generally more deterministic compared to the three types of lexical ambiguities.
This can be attributed to that modern neural networks (NNs) lack uncertainty awareness and are prone to overconfidence when making decisions~\cite{guo2017calibration}, which results in LMs understanding one potential interpretation and respond accordingly without seeking further clarifications.
Furthermore, semantic ambiguity has a more significant impact on output lengths than structural and syntactic ambiguities.
We speculate that this is because some words with multiple meanings such as \textit{``trunk''} or \textit{``monitor''} are more likely to cause confusions for a LM and thereby triggering longer responses, e.g., by enumerating all potential interpretations of a word sense and asking for explanations.

Fig.~\ref{fig:uncertainty_latency}b plots the correlation between inference latency and output length for sentences that contain different types of uncertainties.
We observe that inference latency is proportional to the output length, with longer outputs generally requiring larger inference latencies.
Some sentences with uncertainties such as open-endedness and multi-partness may even take over 700ms for a LM to generate corresponding responses, which is 2$\sim$4 times the latency of normal sentences. 
This presents a substantial opportunity for system-level optimization, as resource manager can leverage this uncertainty impact as an estimation of task execution times to enhance system efficiency and resource utilization.

\subsection{Predicting the Output Length for a Given Input}

\label{sec:uncertainty_prediction}

\begin{lstlisting}[float=t, language=Python, label={lst:ambiguity-rules}, caption={Code for measuring vague expression scores.}]
def vague_expressions_score(sentence, weight=1):
    vague_count = 0
    words = word_tokenize(sentence)
    stem_words = [lemmatizer.lemmatize(word) for word in words]
    # VAGUE_WORDS are pre-defined in the literature
    for phrase in VAGUE_WORDS:
        matches = stem_words.count(lemmatizer.lemmatize(phrase))
        vague_count += matches
    return weight * vague_count
\end{lstlisting}

Upon observing that inference latency is determined by output length, we develop methods that can accurately yet efficiently predict such length for a given input at runtime. 
As discussed earlier, uncertainty of an input text may increase the output length and thus negatively impact inference latency. Our methods shall take uncertainty into account when making such predictions.

\begin{minipage}{0.45\textwidth}
\begin{shaded}
    \noindent \textbf{Uncertainty score}: In this work, we define uncertainty score for an input text as the estimated number of tokens (output length) required to formulate a comprehensive and unambiguous response that sufficiently addresses the posed inquiry.
\end{shaded}
\end{minipage}

\noindent\textbf{Input length.}
Intuitively, longer inputs may lead to LMs generating longer outputs, even without considering uncertainty.
We demonstrate the impact of this naive heuristic on output lengths in Fig.~\ref{fig:rule-length}a.
We observe that although the correlation is not deterministic and noisy, longer input lengths generally induce longer generated outputs.
This inspires us to further improve it by considering uncertainty.

\noindent\textbf{Single rule.}
We measure the intensity of each uncertainty using hand-crafted rules introduced in the literature.
Specially, we use the spaCy\footnote{\url{https://spacy.io/}} language tool to tokenize input text and obtain the Part-of-Speech~(PoS) tag for each token in the original text.
Then, we quantify uncertainty scores by searching for pre-defined patterns inherently existing in each uncertainty source using regular expressions. 
Listing~\ref{lst:ambiguity-rules} shows an example code for quantifying vague expression uncertainty.
Note that for input sentences that do not contain the defined six uncertainty sources, we use input lengths as their single rule scores.
We evaluate the correlation between single rule scores and the output lengths for inputs containing the corresponding type of uncertainty in Fig.~\ref{fig:rule-length}b.
We observe that the correlation is slightly more apparent and less noisy, which demonstrates the impact of uncertainty on LM generation process.

\begin{figure*}
    \centering
    \includegraphics[width=\textwidth]{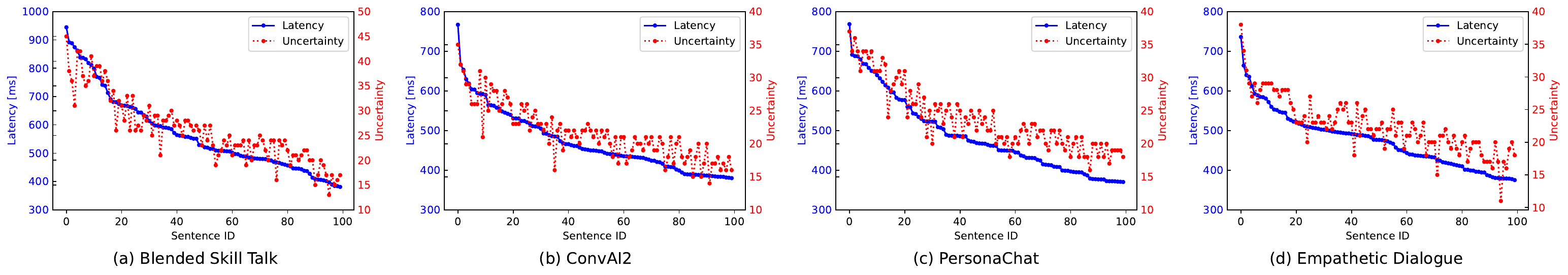}
    \caption{Distribution of latency and corresponding uncertainty on four benchmark DG datasets, (a)~\textit{Blended Skill Talk}, (b)~\textit{ConvAI2}, (c)~\textit{PersonaChat}, (d)~\textit{Empathetic Dialogue}. 
    The data points are ranked by descending order of latency.}
    \label{fig:uncertainty-latency-dg}
    \vspace{-0.3cm}
\end{figure*}

\noindent\textbf{Weighted rule.}
The previous method assumes a primary uncertainty source for each sentence, which is not generic for real-world test cases that may contain multiple uncertainty sources.
Instead, we measure the six defined uncertainty scores for a given text and assign a weight to each category by learning a linear regression to the previously fitted line.
We evaluate the correlation between weighted rule scores and output lengths for inputs with the corresponding type of uncertainty in Fig.~\ref{fig:rule-length}c.
We observe that the dependency between uncertainty scores and output lengths noticeably increases, without more data points getting close to the trend line.

\noindent\textbf{Lightweight model.}
While hand-crafted rules can capture certain uncertainty for sentences, they are heuristic methods and not comprehensive enough since the data distribution is not learned.
To make such estimation more reliable, we introduce a data-driven black-box lightweight (LW) multi-layer perceptron (MLP)~\cite{ramchoun2016multilayer} that takes the six rule-based scores as features and predicts the output length for any given query.
Specifically, we train a LW model on the training sets of four benchmark datasets and evaluate the correlation between its predictions and output lengths for unseen queries in the test sets in Fig.~\ref{fig:rule-length}d.
We observe the output lengths are almost linearly dependent on our predicted scores, with only few noisy samples.
We further evaluate the correlation between the predicted uncertainty scores and averaged inference latency across different LMs on the four benchmark datasets in
Fig.~\ref{fig:uncertainty-latency-dg}.
The predicted scores are highly consistent with the inference latencies across all datasets, i.e., sentences with smaller uncertainties generally require larger inference latencies.
This suggests that our method can precisely estimate LM execution times for any unseen query in real-world dialogue scenarios.

\subsection{System-level Optimization Opportunities}

We now illustrate several precious system-level optimization ideas enabled by leveraging uncertainty score metric.

\begin{figure}
    \centering
    \includegraphics[width=0.48\textwidth]{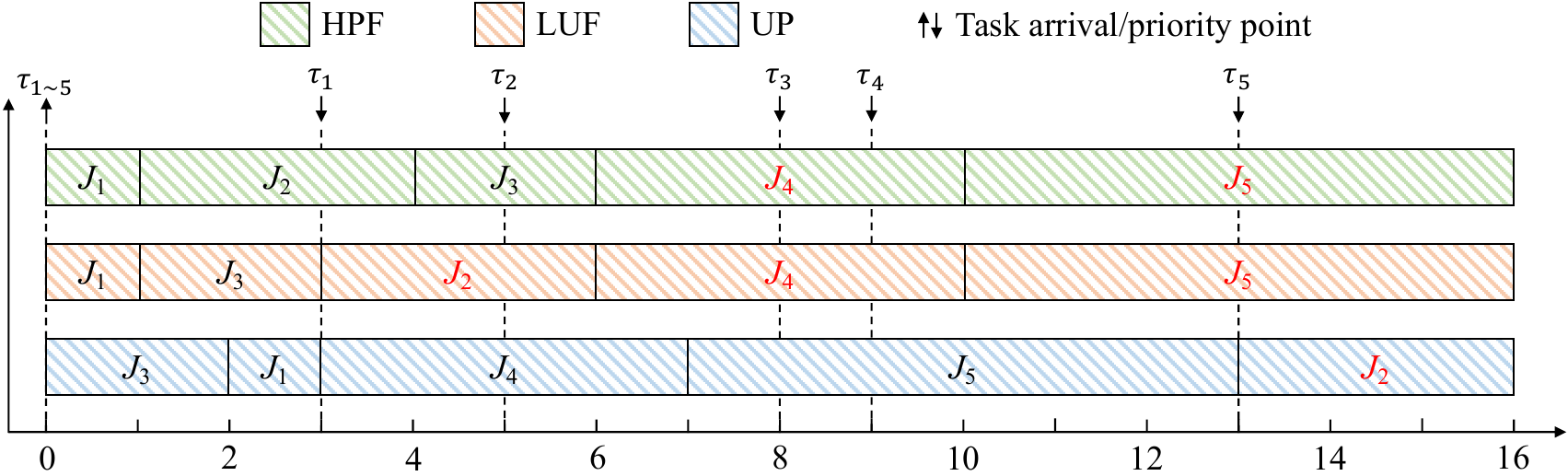}
    \caption{Prioritization example for HPF, LUF, and UP.
    $J_i$ denotes the $i$-th task, $\tau_i$ denotes its arrival/priority point. Tasks depicted in red color denote those missing their priority points.
    }
    \label{fig:prioritization_example}
    \vspace{-0.3cm}
\end{figure}

\noindent\textbf{Prioritization.} 
Online queries, though without intrinsic deadlines, have \textit{priorities} (e.g., urgency of the task) that can be specified by \Approach{} using the priority point parameter according to their estimated workloads.
Leveraging the uncertainty score of each task (i.e., the estimated number of output tokens of each input), the scheduler shall make better prioritization decisions.
Intuitively, prioritizing tasks that require shorter execution times and earlier priority points would improve throughput and timing correctness (often due to reduced head-of-line blocking), as illustrated in Fig.~\ref{fig:prioritization_example}. 
In this example, five tasks that arrive at the same time (the length of each block presents its execution time) are scheduled by three strategies, namely Highest Priority Point First (HPF), Least Uncertainty First (LUF), and \Approach{} utilizing Uncertainty-aware Prioritization (UP).
As a result, HPF and LUF respectively miss two ($J_4$ and $J_5$) and three ($J_2$, $J_4$, and $J_5$) priority points, whereas UP misses only one priority point ($J_2$).

\begin{figure}
    \centering
    \includegraphics[width=0.49\textwidth]{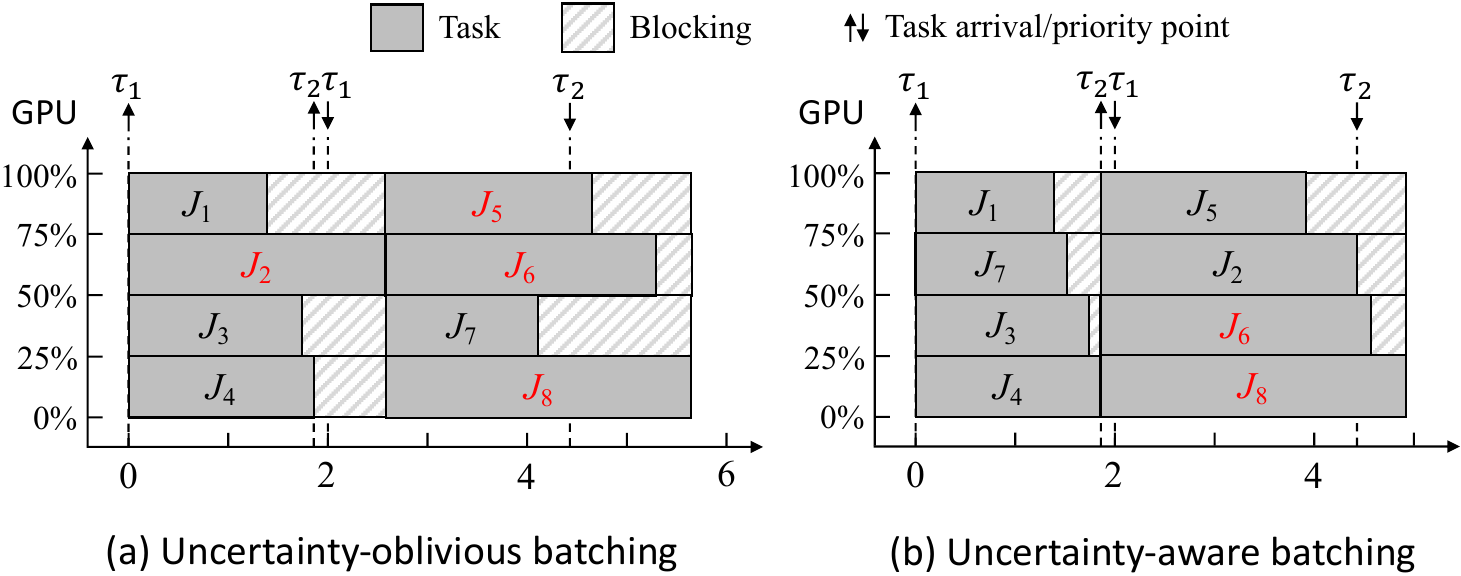}
    \caption{Comparison of (a)~random batching and (b)~consolidation using uncertainty on eight tasks with a batch size of four. $\tau_i$ denotes the arrival/priority point for the $i$-th batch. Tasks with red notations miss the priority points.}
    \label{fig:consolidate}
    \vspace{-0.3cm}
\end{figure}

\noindent\textbf{Consolidation.}
In any heavily-loaded systems requiring machine learning workload multitasking, batch execution is a commonly-used method to enhance response time and timing correctness.
Our estimated uncertainty scores can assist deciding which tasks shall be batched and executed together to better utilize hardware resources.
Fig.~\ref{fig:consolidate} describes this idea using an intuitive example comparing two batch executions for eight tasks with a batch size of four.
Fig.~\ref{fig:consolidate}a presents a schedule under uncertainty-oblivious batching, e.g., HPF where tasks in each batch have similar priority points. 
Four tasks ($J_2$, $J_5$, $J_6$, $J_8$) miss priority points with a fairly low GPU utilization.
Fig.~\ref{fig:consolidate}b describes uncertainty-aware batching, where tasks in each batch have similar uncertainty scores.
Only two tasks ($J_6$, $J_8$) miss priority points with an improved GPU utilization and shorter response time.

\begin{figure}
    \centering
    \includegraphics[width=0.47\textwidth]{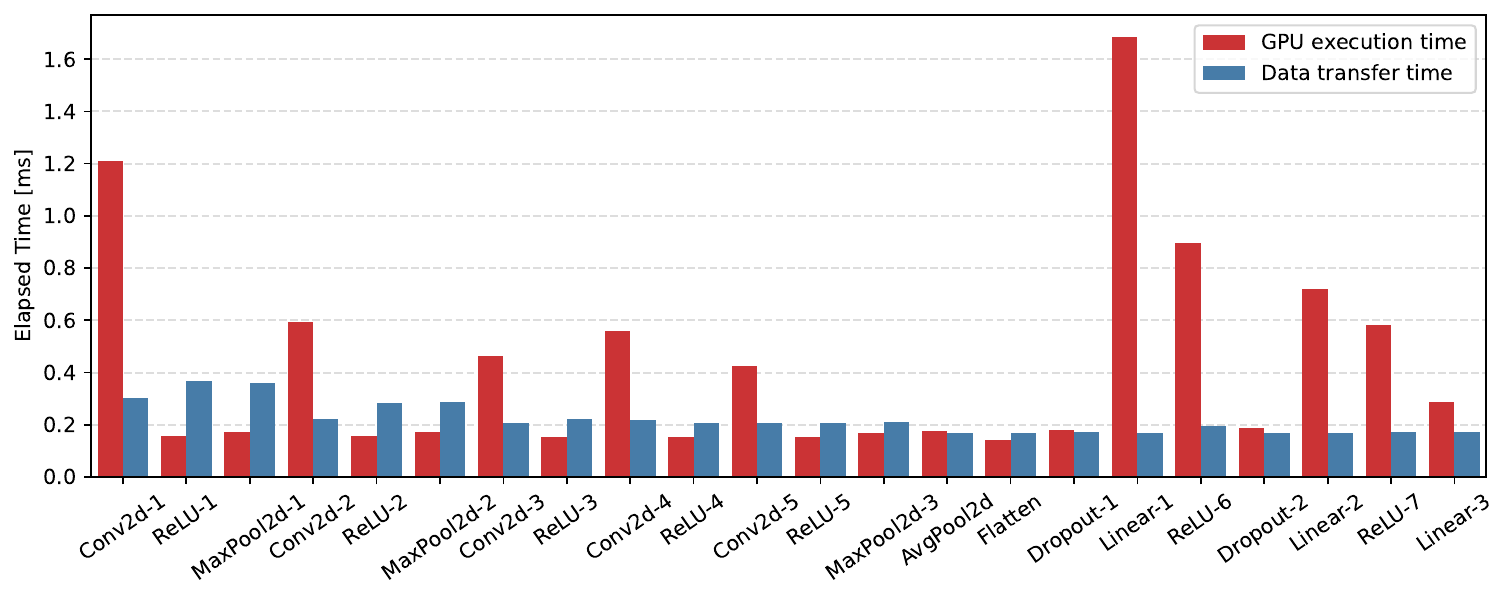}
    \caption{Data transfer time (offloading) compared to GPU execution time for AlexNet.}
    \label{fig:offloading_alexnet}
    \vspace{-0.3cm}
\end{figure}

\noindent\textbf{Strategic offloading to CPU.}
Previous works~\cite{DBLP:conf/rtss/KangLLSC21,DBLP:conf/rtss/JiYKASDK22} and our experiments indicate that offloading machine learning workloads to CPU cores often introduces non-negligible communication and synchronization overhead, negating the benefits of parallel utilization of both CPUs and GPUs.
Fig.~\ref{fig:offloading_alexnet} depicts an illustrative example, where we compare the layer-wise data transfer cost with layer-wise GPU execution times for running AlexNet~\cite{AlexNet}. 
As seen, 
data transfer takes nearly the same amount of time as GPU execution for the majority of layers.
Nonetheless, under overloaded situations or scenarios containing computation-demanding workloads, \Approach{} could identify such tasks by checking whether the estimated uncertainty scores exceeds a pre-defined threshold. 
The scheduler can then decide whether offloading such demanding tasks to CPUs can improve the overall efficiency of the system. 
While it is likely that the negative impact due to offloading and communication can be totally negated by freeing up the precious GPU resource for executing other normal tasks, our intuition is to leverage uncertainty scores to reflect different levels of task demand and offload demanding tasks to CPU.
This strategic offloading balances the workload between CPU and GPU, enabling efficient use of system resources and ensuring that the overall system remains responsive and productive.

\section{Design of \Approach{}}

\subsection{Design Overview}

\begin{figure}[]
    \centering
    \includegraphics[width=0.42\textwidth]{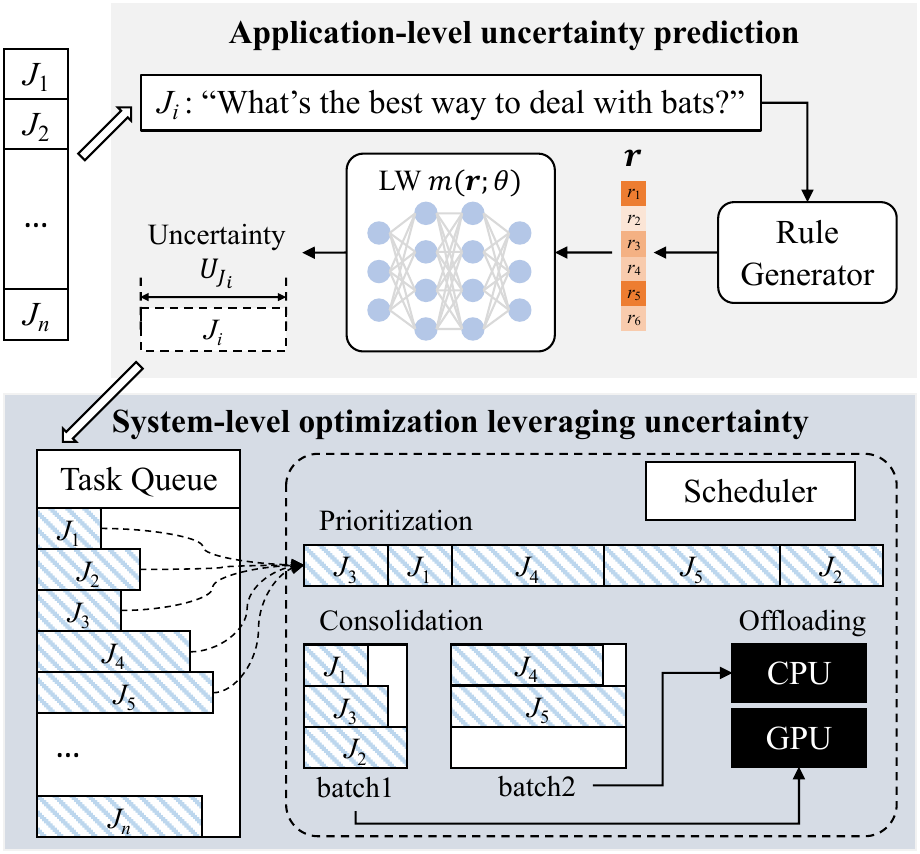}
    \caption{Design overview of \Approach{}. 
    }
    \label{fig:architecture}
    \vspace{-0.3cm}
\end{figure}

In this section, we illustrate the overall design of \Approach{}, as shown in Fig.~\ref{fig:architecture}.
\Approach{} comprises two major components: an application-level framework that quantifies task uncertainty, and a system-level framework that leverages this information for optimized scheduling (prioritization, dynamic consolidation) and resource allocation (strategic offloading).

Defined in Sec.~\ref{sec:uncertainty_prediction}, the uncertainty score of an input text reflects the required output length and thus, its execution times.
Leveraging this critical uncertainty information of input texts, \Approach{} develops an uncertainty-aware system-level resource manager that makes better scheduling decisions. 
To ensure timing correctness, \Approach{} introduces an uncertainty-aware priority scheduler that takes into account both uncertainty scores and priority points of tasks to reflect how critical a task is.
By smartly considering both factors, \Approach{} is capable of improving the system's throughput.
Moreover, \Approach{} includes a runtime consolidation mechanism to enhance the system's latency performance through uncertainty-aware batching.
Our uncertainty estimation aids in deciding which tasks should be batched together to better utilize hardware resources.
The system dynamically forms batches of tasks with similar execution times by reordering the tasks in two adjacent batches according to their uncertainty scores.
In this way, tasks within each batch have both similar criticality and latency, leading to improved GPU utilization and less response time.
Lastly, \Approach{} integrates strategic CPU offloading to handle highly-demanding or malicious workloads.
By leveraging uncertainty scores to indicate the demand of a task, \Approach{} strategically offload tasks that may potentially lead to overloaded situations on GPUs to the CPU core to maintain a balanced workload distribution across the system.

\subsection{Uncertainty-aware Prioritization}
\label{sec:prioritization}

For any given input $J$, our rule generator $\textsc{RuleGen}(\cdot)$ first yields a feature vector containing the intensity of the six linguistic uncertainties.
Then our LW model $m_{\theta}$ takes the feature vector and predicts the final uncertainty score:

\begin{equation}
    \label{eq:uncertainty}
   {u_{J} = m_{\theta}(\textsc{RuleGen}(J))}
\end{equation}

In some scenarios such as conversational AI in healthcare~\cite{bharti2020medbot}, if an LM request has a user-specified deadline $t_J$, \Approach{} can specify the priority point parameter using that deadline ($d_J$ in Eq.~\ref{eq:priority} is replaced by $t_J$); whereas most LM-assisted dialogue systems do not have such user-specified deadlines.
Based on our observations in Fig.~\ref{fig:rule-length}e where longer inputs generally induce longer outputs, we empirically define a priority point for each task according to its input length $d_J = \varphi_{f} |J|$, where $\varphi_f$ is a coefficient that projects input length to the latency of an LM $f$.

A straightforward way of factoring in both uncertainty and priority point into a system is to use the concept of ``slack'' $(\zeta)$, which measures the remaining time until the priority point:

\begin{equation}
    p_{J} = \frac{1}{\zeta_{J}} = \frac{1}{d_{J} - r_{J} -\eta_{f} \cdot u_{J}}
\end{equation}

Here $r_{J}$, $d_{J}$ denote the arrival time and priority point of the task, respectively.
The term $u_{J}$ presents the uncertainty score of the task, reflecting the estimated output length, while $\eta_{f}$ is a coefficient that projects output lengths to latencies, regarding the LM $f$. 
This slack-based approach prioritizes urgent tasks that are close to their priority points, which is suitable for systems with stringent priority point constraints and relatively stable task execution times.

However, for on-device LM systems facing workloads with high variability in uncertainties, such as input texts with a large uncertainty range causing LMs to generate outputs with varied lengths, a more flexible approach that can prioritize tasks with shorter execution times when needed ensures more predictable and consistent system performance.
In \Approach{}, we design Uncertainty-aware Prioritization (UP) where each task is assigned a priority $p_{J}$ that reflects its weighted criticality:

\begin{equation}
\label{eq:priority}
    p_{J} = \frac{1-\alpha \cdot u_{J}}{d_{J}-r_{J}-\eta_{f} \cdot u_{J}}
\end{equation}

Here $\alpha$ is a system-level hyper-parameter that provides a control over the impact of uncertainty on the priority.
Specifically, $d_{J}-r_{J}-\eta_{f} \cdot u_{J}$ represents the estimated slack for the execution of the task, and $\alpha \cdot u_{J}$ is a scaled uncertainty score.
The fraction computes the estimated execution time after considering the scaled uncertainty, normalized by how much time is left, to represent the criticality of a task.
The intuition behind this priority assignment is that a task with a shorter slack window or smaller uncertainty score should have a higher priority.
This ensures that tasks with imminent priority points or short execution times are attended to promptly, enhancing the likelihood of meeting their priority points.
The factor $\alpha$ provides a level of adaptability to the system.
A larger value of $\alpha$ implies that the system is placing a higher emphasis on tasks with lower uncertainties, regardless of how soon their priority points are, while a smaller $\alpha$ value reduces the impact of uncertainty on the priority calculation, placing a higher emphasis on the remaining time until the priority point. 
We search an optimal $\alpha$ value from 0 to 2.0 with a increment of 0.1 by testing the corresponding response time (see Fig.~\ref{fig:param_study}a).

\subsection{Dynamic Consolidation}
\label{sec:consolidation}

In the dynamic consolidation process, we aim to enhance the overall system efficiency by executing batches of tasks with similar estimated uncertainties, as they are more likely to have comparable processing requirements. 
The intuition is that executing tasks with similar workload characteristics as a batch can potentially lead to better resource utilization and reduced overheads, as illustrated in Fig.~\ref{fig:consolidate}.
Specifically, we maintain a queue of tasks sorted by the priority based on our UP algorithm (Eq.~\ref{eq:priority}).
We then group tasks with similar uncertainty scores together by introducing two hyper-parameters, $\lambda$ and $b$.
Among them, $b$ determines the number of tasks to consider for a batch.
Given a pre-defined batch size $C$, once the current batch accumulates $b \times C$ tasks from the task queue, we reorder these tasks according to their uncertainty scores.
We then select the top-$C$ tasks from this reordered list for execution.
This mechanism ensures that tasks are executed in an order that prioritizes higher urgency as well as shorter execution times.
Additionally, parameter $\lambda$ controls the maximum allowable ratio in uncertainty scores between tasks within a batch. 
As we traverse the sorted list of tasks within the current batch, if we encounter a situation where the uncertainty score of the current task is more than $\lambda$ times that of the previous one, we segment the list at this point.
The tasks preceding this point are executed as a batch, while the remaining tasks are returned to the queue for future processing.
The whole consolidation process unfolds as follows:

\begin{itemize}[leftmargin=*]
    \item Maintain a queue of tasks ordered by descending priority, based on the UP algorithm.

    \item Once accumulating $b \times C$ tasks in the current batch, reorder them in accordance with their uncertainty scores.

    \item Traverse the reordered batch of tasks.  If the uncertainty of a task exceeds $\lambda$ times the uncertainty of the previous task, or if the batch size $C$ is met, segment the list at this point.

    \item Execute the tasks before the segmentation point as a batch, while returning the remaining tasks to the queue.
    
\end{itemize}

Dynamic consolidation provides flexibility in adjusting to varied workload characteristics and system conditions through the adjustment of the parameters $b$ and $\lambda$.
For instance, in scenarios where tasks exhibit diverse uncertainty scores, a smaller $b$ or larger $\lambda$ can be utilized to ensure that only tasks with similar uncertainties are grouped together. 
Conversely, if tasks have similar uncertainty scores, a larger $b$ or smaller $\lambda$ will form larger batches, potentially achieving higher system throughput.
Moreover, dynamic consolidation can help balance the trade-off between throughput and predictability. 
By executing tasks with similar uncertainties as a batch, the system may exhibit more predictable behaviors, as estimating the execution time of a batch is often simpler than predicting individual task execution times. 
Meanwhile, by executing tasks in batches, the system can potentially achieve higher throughput compared to executing tasks individually.

\subsection{Strategic Offloading to CPU}
\label{sec:offloading}

In the dynamic consolidation process described above, tasks are assigned to batches and then executed based on uncertainty scores.
However, such a process can lead to the situation where some tasks with high uncertainty scores (e.g., malicious, adversarial tasks) may potentially delay the execution of the whole batch, negatively affecting the overall system performance.
To address this, we propose a protective mechanism, termed `strategic offloading', to offload potentially malicious tasks and execute them separately on CPU cores.

In our implementation, we define a parameter $k$ ($0 < k < 1$) which denotes the top-$k$ percentage of uncertainty scores in the training set to control the malicious threshold $\tau$:

\begin{equation}
\label{eq:adv_threhsold}
    \tau = \text{quantile}_{k} \left( \{ m_{\theta}(\textsc{RuleGen}(J)) | J \in \mathcal{D}_{train} \} \right)
\end{equation}

\begin{algorithm}[t]
\caption{Uncertainty-aware framework of \Approach{}}
\label{alg:main_method}
\begin{algorithmic}[1]
\State \textbf{function} \textsc{UaSched}($\alpha$, $\lambda$, $k$, $b$);
\State Initialize a lightweight regressor $m_{\theta}$;
\For {each $J \in \mathcal{D}_{train}$} \Comment{Offline profiling}
\State Rule scores $\mathbf{r}_{J} \leftarrow \textsc{RuleGen}(J)$;
\State LM output $y_{J} \leftarrow f(J)$ with length $|y_{J}|$;
\State Minimize $\mathcal{L}_{MSE} \leftarrow \sum \left \{m_{\theta}(\mathbf{r}_{J}) - |y_{J}| \right \}^2$ and update the parameters of $m_{\theta}$;
\State Record the GPU util. under the current batch size; 
\EndFor
\State Obtain the optimal batch size $C_{f}$ for each LM $f(\cdot)$;
\State $\tau \leftarrow \text{quantile}_{k} \left( \{ m_{\theta}(\textsc{RuleGen}(J)) | J \in \mathcal{D}_{train} \} \right)$;
\For {each $(J, r_{J}, d_{J}) \in \mathcal{D}_{test}$}
\Comment{Online scheduling}
\State Uncertainty score $u_{J} \leftarrow m_{\theta}(\textsc{RuleGen}(J))$;
\State Uncertainty-aware priority $p_{J} \leftarrow \frac{1-\alpha \cdot u_{J}}{d_{J}-r_{J}-\eta_{f} \cdot u_{J}}$;
\State Put $(p_{J}, u_{J}, J, r_{J}, d_{J})$ in the task queue $Q$;
\EndFor
\For {each $(p_{J}, u_{J}, J, r_{J}, d_{J}) \in Q$ in descending $p$ order}
\If {$u_{J} > \tau$}
\State Offload $J$ into the CPU batch; \Comment{Offloading}
\Else
\State Put $J$ into a tmp batch $\mathcal{T}$;
\EndIf
\If {$|\mathcal{T}| = \left \lfloor  b\cdot C_f \right \rfloor $} \Comment{Consolidation}
\State Sort $J \in \mathcal{T}$ in ascending uncertainty $u$ order; 
\State $u_{prev} \leftarrow \mathcal{T}[0]$; count $\leftarrow 0$;
\While {$u_{J} \leq \lambda\cdot u_{prev}$ $\vee$ $\text{count} < C_f$} 
\State $u_{prev} \leftarrow u_J$; count++;
\EndWhile
\State Append $\mathcal{T}[:\text{count}]$ into the GPU batch; 
\State Put $\mathcal{T}[\text{count}:]$ back into the task queue $Q$;
\State Clear the tmp batch $\mathcal{T}$.
\EndIf
\EndFor
\end{algorithmic}
\end{algorithm}

In essence, $\tau$ corresponds to the boundary of the highest $k$-percentile of uncertainty scores.
If the uncertainty score of a task is larger than $\tau$, it is offloaded to a CPU batch for separate execution.
Otherwise, it is assigned to a GPU batch for grouped execution.
Furthermore, we ensure that there is always a batch of tasks ready for execution. 
If the task queue is empty and there are remaining tasks in the GPU batch, these tasks are offloaded for execution. 
Similarly, if there are no tasks in the GPU and CPU batches, the remaining tasks from the task queue are offloaded to the appropriate execution batch based on their uncertainty scores.
This strategic offloading mechanism provides a layer of protection against extreme execution times, ensuring malicious tasks do not excessively delay the execution of a batch and promising a more predictable and reliable system performance, particularly under workloads with high variability. 
By carefully controlling the offloading parameter $k$, this mechanism can be tuned to balance the benefits of grouping tasks for efficient execution against the potential delays caused by malicious tasks.

\subsection{Pseudo Code and Illustration}

Algorithm~\ref{alg:main_method} illustrates the whole framework of \Approach{}, known as \textsc{UaSched}.
It takes several aforementioned control parameters, $\alpha$, $\lambda$, $k$, and $b$, and operates in two main phases: offline profiling and online scheduling.

\begin{table}[t]
  \centering
  \caption{Hardware platforms used in our experiments.}
  \renewcommand\arraystretch{1.3}
  \resizebox{0.4\textwidth}{!}{ 
    \begin{tabular}{c|c|c}
    \hline 
     & {\textbf{Edge Server}} & {\textbf{NVIDIA AGX Xavier}} \\
    \hline
    {\multirow{3}{*}{CPU}}  & 96-core AMD  & 8-core NVIDIA  \\
    & EPYC 7352 & Carmel Armv8.2  \\
    & 24-Core Processor & 64-bit CPU \\
    \hline
    GPU & NVIDIA RTX A4500 & NVIDIA Volta GPU \\
    \hline
    Memory & 512GB & 16GB LPDDR4x \\
    \hline
    Storage & 8TB SSD & 32GB eMMC \\
    \hline
    \end{tabular}
}
  \label{tab:hardware}
\end{table}

\begin{figure}
    \centering
    \includegraphics[width=0.48\textwidth]{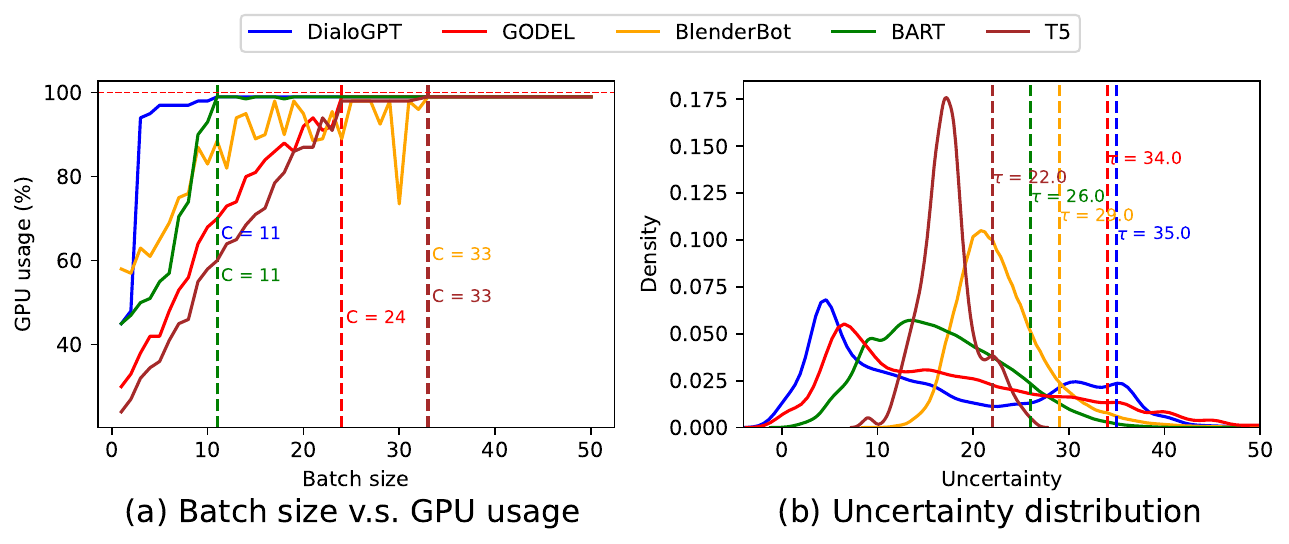}
    \caption{Offline decisions on (a)~optimal batch size $C$ and (b)~malicious threshold $\tau$ ($k=0.9$) for the five LMs.}
    \label{fig:offline}
    \vspace{-0.3cm}
\end{figure}

\noindent\textbf{Offline profiling.} 
The algorithm starts by initializing a LW regressor $m_{\theta}$.
For each task in the training set, $\textsc{RuleGen}(\cdot)$ generates rule scores $\mathbf{r}_{J}$, which is taken by $m_{\theta}$ as features and calculates the output length from the LM.
The algorithm then minimizes the Mean Squared Error (MSE) between the estimated output lengths and the LM output lengths, thereby updating the LW model.
It also records GPU utilization to determine the minimum batch size $C_f$ for the LM $f(\cdot)$ that can better utilize hardware resources, e.g., when GPU usage reaches 100\%.
Finally, it determines the malicious threshold $\tau$ according to the uncertainty score distribution.

\begin{figure*}
  \centering
  \begin{minipage}[b]{0.72\textwidth}
    \includegraphics[width=\textwidth]{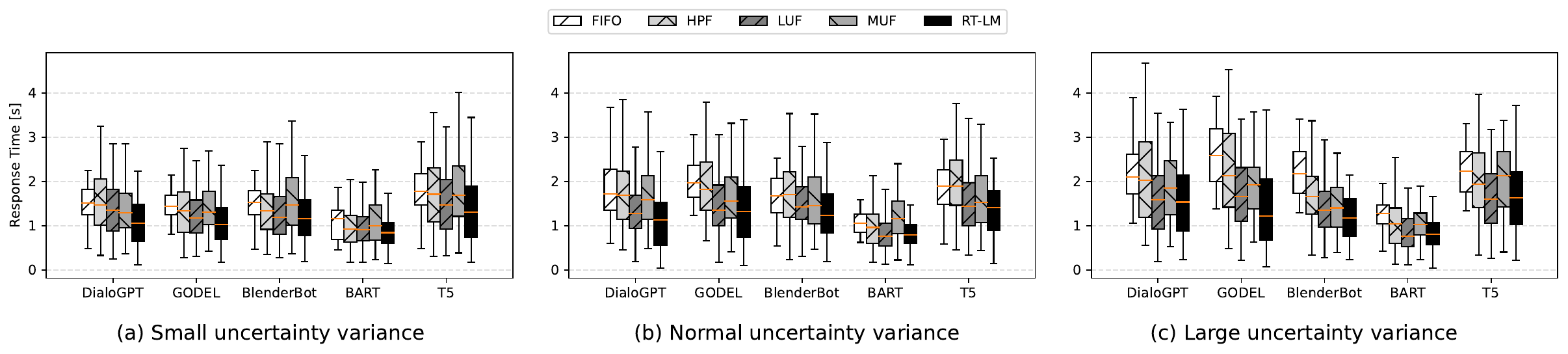}
    \caption{The distribution of response time across five LMs for sentences with (a)~small, (b)~normal, and (c)~large uncertainty variance on the edge server.}
    \label{fig:response_server}
  \end{minipage}
  \hfill
  \begin{minipage}[b]{0.25\textwidth}
    \includegraphics[width=\textwidth]{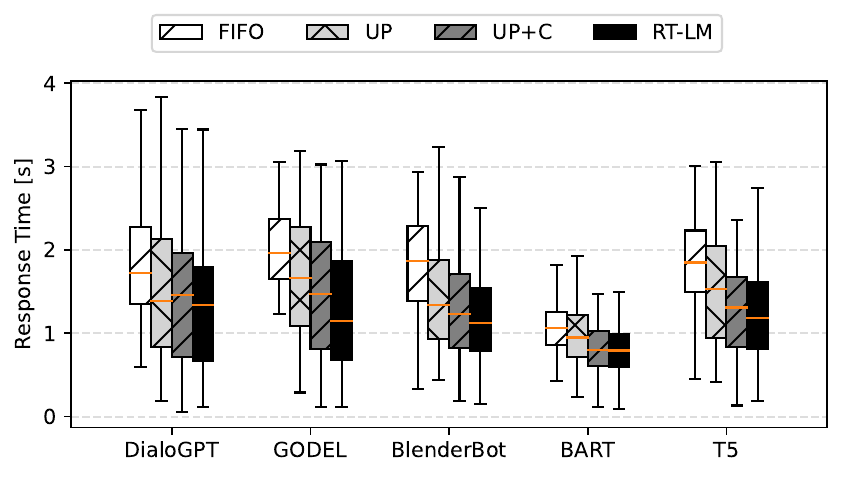}
    \caption{Ablation study of response time on the edge server.
    }
    \label{fig:ablation_server}
  \end{minipage}
  \vspace{-0.3cm}
\end{figure*}

\begin{figure*}
  \centering
  \begin{minipage}[b]{0.72\textwidth}
    \includegraphics[width=\textwidth]{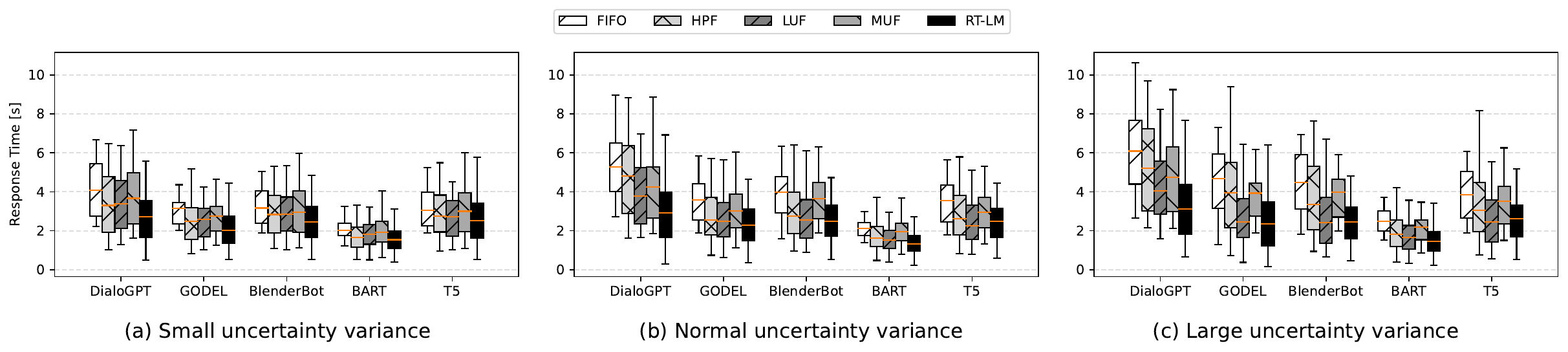}
    \caption{The distribution of response time across five LMs for sentences with (a)~small, (b)~normal, and (c)~large uncertainty variance on the AGX Xavier.}
    \label{fig:response}
  \end{minipage}
  \hfill
  \begin{minipage}[b]{0.25\textwidth}
    \includegraphics[width=\textwidth]{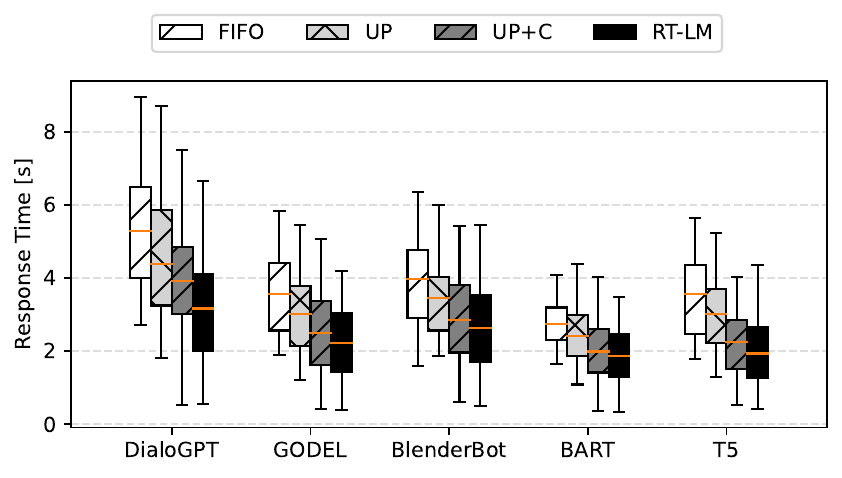}
    \caption{Ablation study of response time on the AGX Xavier.
    }
    \label{fig:ablation}
  \end{minipage}
  \vspace{-0.3cm}
\end{figure*}

\noindent\textbf{Online scheduling.}
The algorithm iterates over tasks in the test set, calculating uncertainty scores using the pre-trained LW model $m_{\theta}$, and then placing them into a task queue.
The tasks are then popped and processed in a descending order of priority scores.
If a task's uncertainty score is greater than the threshold, it is offloaded to a CPU batch; otherwise, it is placed in a temporary batch.
If the temporary batch reaches a size of $b \cdot C_f$, the scheduler sorts tasks in the batch in ascending order of uncertainty scores. 
It then segments the batch at a point where the current uncertainty score is larger than $\lambda$ times that of the previous one or if the pre-defined batch size $C_f$ has been reached. 
The segmented tasks are offloaded to a GPU batch, while the remaining ones are put back into the queue.

\section{Implementation and Evaluation}

\subsection{Experiment Setup}

\noindent\textbf{Testbeds.}
We implement \Approach{} and conduct an extensive set of experiments on an edge server, as shown in Table~\ref{tab:hardware}, simulating the single-device multitasking scenarios of online chatbots or services, and live-translation services.

\noindent\textbf{Benchmark.}
We evaluate \Approach{} across five state-of-the-art LMs that are widely used in dialogue systems---DialoGPT~\cite{dialogpt}, GODEL~\cite{GODEL}, BlenderBot~\cite{blenderbot}, BART~\cite{lewis-etal-2020-bart}, and T5~\cite{T5}---on four benchmark datasets: \textit{Blended Skill Talk}~\cite{bst}, \textit{PersonaChat}~\cite{personachat}, \textit{ConvAI2}~\cite{convai2}, and \textit{Empathetic Dialogues}~\cite{empathetic_dialogue}.
We use the pre-trained versions of these models---\textit{DialoGPT-medium}, \textit{GODEL-v1\_1-base-seq2seq}, \textit{blenderbot-400M-distill}, \textit{bart-base}, \textit{t5-base} and annotated datasets released by Hugging Face\footnote{\url{https://huggingface.co/}}.

\noindent\textbf{Metrics.}
We evaluate \Approach{}'s performance w.r.t. the average response time, throughput, and runtime overhead.  
We also delve deeper into the effect of different components of \Approach{} on the system-level performance, the robustness of \Approach{} against different parameter settings, and its effectiveness under different proportions of malicious tasks.

\noindent\textbf{Hyper-parameters.}
For the offline profiling, we initialize a lightweight MLP which has four layers of hidden size [100, 200, 200, 100], and train the model with a learning rate of 1e-4.
We record the average GPU usage for the five LMs with different batch sizes in Fig.~\ref{fig:offline}a.
Specifically, we choose an optimal batch size (i.e., minimum batch size that a LM can reach 100\% GPU usage) of 11, 24, 33, 11, 33, for DialoGPT, GODEL, BlenderBot, BART, T5, respectively.
We further record the distribution of uncertainty scores for each LM in Fig.~\ref{fig:offline}b, and select a malicious threshold of 35, 34, 29, 26, 22 for DialoGPT, GODEL, BlenderBot, BART, T5, respectively.
We set the uncertainty-weight $\alpha$ as 1.0, the output-latency coefficients $\eta$ as 0.05, 0.04, 0.1, 0.05, 0.04, and the input-latency coefficients $\varphi$ as 0.08, 0.10, 0.13, 0.08, 0.07 across the five LMs for priority assignment; 
$\lambda$, $b$ as 1.5, 1.8, respectively for dynamic consolidation; and $k$ as 0.9 for protective mechanism.
To gather necessary statistics, we employ the tegrastats utility for recording GPU and CPU memory usage.
Additionally, we use Python's time library to track the arrival and end time of each task, as well as the latency incurred by \Approach{}.

\begin{table*}[]
    \centering
    \caption{Maximum response time (s) and percentage of improvement for sentences with small, normal, and large uncertainty variance on the edge server. The evaluated methods consist of uncertainty-oblivious (former) and uncertainty-aware (latter) ones. \textbf{Bold} numbers denote the best metric values among them.}
    \resizebox{0.99\textwidth}{!}{
    \begin{tabular}{c|ccc|ccc|ccc|ccc|ccc}
    \toprule
    \multirow{2}{*}{Method} & \multicolumn{3}{c|}{DialoGPT} & \multicolumn{3}{c|}{GODEL} & \multicolumn{3}{c|}{BlenderBot} & \multicolumn{3}{c|}{BART} & \multicolumn{3}{c}{T5} \\
    \cmidrule{2-16}
    & Small & Normal & Large & Small & Normal & Large & Small & Normal & Large & Small & Normal & Large & Small & Normal & Large \\
    \midrule
    FIFO & 2.25 & 3.75 & 3.90 & \textbf{2.15} & 3.06 & 3.93 & \textbf{2.24} & 2.52 & 3.41 & \textbf{1.87} & 1.93 & 1.95 & \textbf{2.90} & 2.95 & 3.30 \\
    HPF & 3.25 & 3.92 & 4.68 & 2.75 & 3.79 & 4.53 & 2.90 & 3.54 & 3.37 & 2.34 & 2.13 & 2.63 & 3.56 & 4.13 & 3.97 \\
    \midrule
    LUF & 2.85 & \textbf{2.77} & 3.55 & 2.47 & 3.06 & 3.41 & 2.86 & 2.79 & 2.93 & 1.98 & 1.82 & 2.19 & 3.24 & 3.43 & \textbf{3.17} \\
    MUF & 3.03 & 3.68 & 3.93 & 3.52 & 3.74 & 4.21 & 3.36 & 3.52 & 3.10 & 2.97 & 3.00 & 2.38 & 4.01 & 3.30 & 3.98 \\
    \multirow{2}{*}{\Approach{}} & \textbf{2.24} & 2.96 & \textbf{3.18} & 2.52 & \textbf{2.80} & \textbf{3.17} & 2.92 & \textbf{2.26} & \textbf{2.38} & 1.93 & \textbf{1.66} & \textbf{1.86} & 3.45 & \textbf{2.64} & 3.25 \\
    & -0.4\% & -21.1\% & -18.5\% & +17.2\% & -8.5\% & -19.3\% & +30.4\% & -10.3\% & -30.2\% & +3.2\% & -14.0\% & -4.6\% & +19.0\% & -23.4\% & -1.5\% \\
    \bottomrule
    \end{tabular}}
    \label{tab:max_response}
    \vspace{-0.3cm}
\end{table*}

\noindent\textbf{Workload setup.} 
Real-world human-generated processes, such as phone calls to a call center, can often be represented as a Poisson process, where the number of arrivals within a specific time interval is governed by a Poisson distribution~\cite{cinlar2013stochastic,ross2014probability}. 
Given the independent nature of user queries in our context, we adopt a similar model to simulate task arrivals. 
This model is principally defined by its average arrival rate, denoted as $\beta$ (representing queries per minute).
We generated synthetic traces by sampling inter-arrival times from an exponential distribution with differing mean $\mu=\frac{1}{\beta}$ to modulate the arrival rate.
To create time-varying synthetic workloads, we continuously evolve the workload generator across different exponential distributions throughout the process.
This involves iterating through integer values of $\beta$ ranging from 10 to 150. 
For each minute, we sample from the corresponding exponential distribution, ensuring a comprehensive representation of workload scenarios, from light-load phases to high-traffic peaks.
Following the generation of these traces, we shuffle the test dataset and map them to the created arrival patterns.
To enhance realism, acknowledging that users may require some time to complete a query, we introduced a wait time interval $\xi=2$ seconds so that tasks arriving within this span are processed as either a single batch or multiple batches\footnote{We conducted supplementary experiments using diverse sets of $\mu$ and $\xi$ values. The findings consistently align with the trends observed in Fig.\ref{fig:response_server}$\sim$\ref{fig:response}.}.

\subsection{Latency Performance}

We evaluate the latency performance of various strategies by calculating their response time -- the time elapsed between a task's end time and its arrival time across the five LMs.
Naturally, a lower average response time indicates a more efficient system.
We compare \Approach{} to the following baselines:

\begin{itemize}[leftmargin=*]
    \item First-In-First-Out (FIFO): Tasks are queued based on their arrival times, creating uncertainty-oblivious random batches with a fixed size for execution.  

    \item Highest Priority-Point First (HPF)~\cite{edf_textbook}: Tasks with higher priority points are prioritized. This approach batches tasks with similar priority points together, maintaining a fixed batch size, yet remains uncertainty-oblivious.

    \item LUF: Tasks with lower uncertainty scores are given precedence. Those with comparable uncertainty scores (or execution times) are batched together using a fixed size. 

    \item Maximum Uncertainty First (MUF): This strategy prioritizes tasks with higher uncertainty scores. Those with analogous scores are batched together with a set size.
    
\end{itemize}

To gauge the impact of uncertainty on system-level performance, we evaluate all methods across three subsets of tasks featuring small, medium, and large variance of uncertainty scores on the edge server.
Fig.~\ref{fig:response_server} demonstrates the distribution of response time values, while Table~\ref{tab:max_response} records the worst-case response time for each method across task subsets. 
From our observations: 
1)~Uncertainty-aware strategies tend to surpass uncertainty-oblivious ones, especially when input data exhibits varied uncertainty scores.
For the small-variance subset, all methods display similar response times in Fig.~\ref{fig:response_server}a, with the maximum values of LUF, MUF, \Approach{} even larger than FIFO, HPF in some cases in Table~\ref{tab:max_response}, but on the large-variance subset, LUF, MUF, \Approach{} consistently outperform FIFO and HPF.
This is because when tasks exhibit similar workloads, all strategies essentially mimic FIFO.
However, when there's significant variance in task uncertainty, grouping tasks with analogous uncertainty scores reduces the likelihood of computation-intensive tasks holding up the entire batch.
2)~Generally, LUF produces a better performance than MUF.
By prioritizing tasks with high uncertainty, MUF can inadvertently cause the entire system to lag, thus compromising average response times.
3)~\Approach{} consistently exhibits superior performance, achieving the most efficient response times across all LMs.
The average response time of \Approach{} is roughly 0.8s less than FIFO for BART in Fig.~\ref{fig:response_server}c; and its maximum response time is up to 30\% smaller than FIFO for BlenderBot in Table~\ref{tab:max_response}.
This suggests that considering both execution times and priority points in task prioritization can further optimize latency performance.
This dual consideration ensures \Approach{} is versatile across varied workload distributions.
4)~Larger LMs are more sensitive to variations in task uncertainty, requiring even more execution times for tasks with high uncertainty scores, thereby benefiting more from uncertainty-aware strategies, e.g., \Approach{} improves the maximum response time over FIFO to a larger extent for GODEL and BlenderBot (20\% and 30\%) than other LMs.

\subsection{Throughput Performance}

\begin{table*}[]
    \centering
    \caption{Average throughput for sentences with small, normal, and large uncertainty variance on the edge server.}
    \resizebox{0.99\textwidth}{!}{
    \begin{tabular}{c|ccc|ccc|ccc|ccc|ccc}
    \toprule
    \multirow{2}{*}{Method} & \multicolumn{3}{c|}{DialoGPT} & \multicolumn{3}{c|}{GODEL} & \multicolumn{3}{c|}{BlenderBot} & \multicolumn{3}{c|}{BART} & \multicolumn{3}{c}{T5} \\
    \cmidrule{2-16}
    & Small & Normal & Large & Small & Normal & Large & Small & Normal & Large & Small & Normal & Large & Small & Normal & Large \\
    \midrule
    FIFO & 21.68 & 18.00 & 15.68 & 17.89 & 18.28 & 13.36  & 21.15 & 17.90 & 17.63 & 32.30 & 30.62 & 25.92 & 17.86 & 16.57 & 16.15 \\
    HPF & 20.26 & 19.27 & 16.41 & 19.55 & 18.69 & 13.99 & 21.26 & 18.28 & 17.32 & \textbf{33.59} & 30.22 & 26.56 & 18.10 & 16.75 & 17.18 \\
    \midrule
    LUF & 23.19 & 21.09 & 19.97 & 19.71 & 19.17 &  17.68 & \textbf{21.34} & 19.48 & 19.81 & 32.86 & 31.02 & 28.52 & 19.75 & 18.94 & 18.84 \\
    MUF & 22.40 & 20.06 & 19.44 & 19.14 & 18.34 & 16.76 & 21.20 & 18.92 & 20.08 & 32.03 & 31.28 & 27.97 & 19.58 & 17.78 & 17.52 \\
    \Approach{} & \textbf{24.61} & \textbf{23.89} & \textbf{22.34} & \textbf{23.73} & \textbf{21.54} & \textbf{19.78} & 21.12 & \textbf{20.66} & \textbf{20.80} & 32.14 & \textbf{31.71} & \textbf{28.64} & \textbf{22.28} & \textbf{21.94} & \textbf{20.03} \\
    \bottomrule
    \end{tabular}}
    \label{tab:throughput}
    \vspace{-0.3cm}
\end{table*}

We further evaluate the throughput of various strategies as the average completed tasks per minute, across the five LMs, on the edge server. 
As expected, a higher throughput implies a more efficient system. Table~\ref{tab:throughput} summarizes the results on the three subsets.
We observe the throughput profiles of all methods are highly consistent with their latency performance metrics. 
Specifically, uncertainty-aware strategies notably exhibit larger advantages over uncertainty-oblivious ones when the uncertainty variance of test inputs grows, e.g., \Approach{} can process over 6 more tasks per minute than FIFO, with DialoGPT in the large-variance subset. 
Among these, LUF is generally superior to MUF. \Approach{}, however, stands out by consistently outperforming all other strategies. 
Moreover, uncertainty-aware strategies, particularly on larger LMs, can significantly boost system efficiency, e.g., \Approach{} boosts the average throughput by 10\% to 30\% for BART and GODEL.

\subsection{Ablation Study}

To elucidate the superiority of \Approach{}, we conduct an ablation study investigating the individual contributions of each component of our method to the response time and throughput performance:

\begin{itemize}[leftmargin=*]
    \item Uncertainty-aware prioritization (UP): We compare uncertainty-oblivious prioritization strategies, namely FIFO and HPF, with UP for response time and throughput evaluation, respectively.
    \item Dynamic consolidation: We contrast UP (using static batching) with its dynamic consolidation counterpart (UP+C).
    \item Strategic offloading: We compared UP+C with \Approach{}, which facilitates execution of malicious tasks on the CPU.
\end{itemize}

Fig.~\ref{fig:ablation_server} illustrates the subtle improvements of each component-enabled method over its component-oblivious counterpart in terms of reduced response times on the edge server, with \Approach{} consistently outperforming the rest.
For example, UP achieves an average response time of 0.2$\sim$0.7s less than FIFO. 
This indicates all three components of \Approach{} are integral to its superior performance.
Notably, the performance boost derived from prioritization and consolidation is typically larger than offloading, e.g., the average response time gap between UP+C and \Approach{} is smaller than other pairs in most cases. 
This suggests that our prioritization and consolidation are more consequential in improving efficiency.
Interestingly, strategic offloading has slightly more significant impact on larger LMs, e.g., \Approach{} reduces the average response time over UP+C to 0.4s for GODEL, while their performance are nearly the same for BART.
This is because computational demanding tasks have larger impact on sophisticated LMs, causing even more severely overloaded systems.

\subsection{On-Device Evaluation}

Emerging embedded devices, augmented with powerful computing capabilities and LM intelligence~\cite{edge_connected_AI}, have the potential to serve as the local central service in future smart homes. These devices may support hundreds of IoT devices, facilitating concurrent multi-user or multi-device (e.g., refrigerator, air conditioner) communications with a single LM, a concept known as connected intelligence. In this context, we delve into the performance evaluation of various methods on an NVIDIA AGX Xavier (see Table~\ref{tab:hardware}), which is widely used in various applications such as autonomous driving~\cite{kato2018autoware,kisavcanin2017deep} and robotics~\cite{popov2022nvradarnet,Duckiebot(DB-J),SparkFun_JetBot,Waveshare_JetBot}, to reflect the feasibility of \Approach{} in on-device multitasking scenarios.

Fig.~\ref{fig:response} showcases the response time of all evaluated methods across three subsets on the AGX Xavier. The observed patterns largely mirror those seen on the edge server. For instance, uncertainty-aware strategies excel, particularly in subsets with diverse uncertainty characteristics. LUF is generally more efficient than MUF, \Approach{} consistently outperforms other baselines across all LMs, and uncertainty-aware strategies derive greater efficiency benefits from larger LMs, such as GODEL. Furthermore, a comparative analysis between the two platforms reveals an interesting insight: high-performance devices, being quicker in execution, tend to display a smaller disparity in performance across different methods compared to embedded devices. This subtly hints at a diminished relative advantage for \Approach{} on more powerful devices.

Fig.~\ref{fig:ablation} depicts the individual contributions of each \Approach{} component, in terms of reduced response time on the embedded device. The findings align with on the edge server: all three components collectively boost its performance, prioritization and consolidation emerge as more influential factors in enhancing efficiency than offloading, and larger LMs generally derive more pronounced benefits from offloading.

\subsection{Parameter Study}

\begin{figure}
    \centering
    \includegraphics[width=0.48\textwidth]{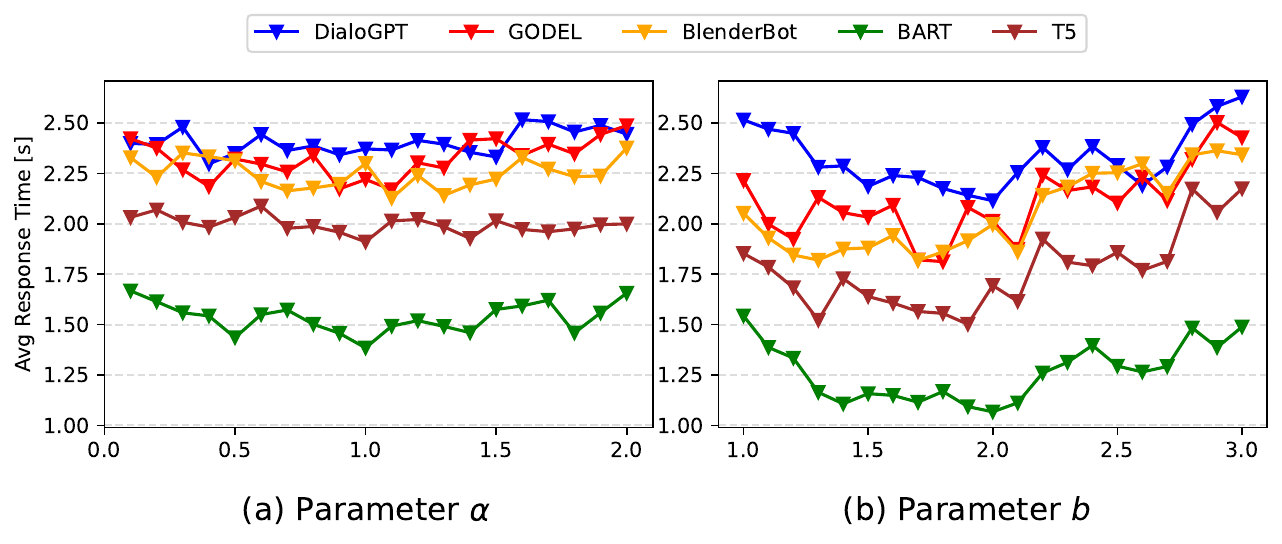}
    \caption{Study of average response time with different values of (a)~$\alpha$ and (b)~$b$ across five LMs on the edge server.}
    \label{fig:param_study}
    \vspace{-0.3cm}
\end{figure}

We explore the impact of two key hyperparameters, $\alpha$ and $b$, which control the influence of uncertainty in priority computation and the batch size determined by the number of tasks, on \Approach{}.
We vary $\alpha$ from 0.1 to 2.0 (with a fixed $b=2.0$) and $b$ from 1.0 to 3.0 (with a fixed $\alpha = 1.0$), incrementing by 0.1 in both cases, and assess the resulting average response time of \Approach{} across different LMs.

Fig.~\ref{fig:param_study}a shows that \Approach{} is robust to changes in $\alpha$, with a maximum divergence in response time not exceeding 0.35s for each LM.
This resilience indicates that UP functions as a well-balanced, uncertainty-aware priority, aptly mediating between priority points and execution times for tasks.
An optimal $\alpha$ value of 1.0 is indicated by our performance metrics. 
Placing a higher emphasis on either uncertainty (larger $\alpha$) or remaining time until the priority point (smaller $\alpha$) results in a slight increase of response time.

Fig.~\ref{fig:param_study}b reveals that $b$ has a more significant impact on latency performance than $\alpha$, with the maximum deviation in response time reaching about 0.75s for T5.
This indicates a considerable dependence of dynamic consolidation on the number of tasks considered for a batch.
Optimal performance is achieved at $b = 1.8$.
Values below or above this introduce inefficiencies, either mimicking static batching or causing delays in task completion due to longer wait time.

\subsection{Evaluating Malicious Scenarios}

\begin{table}[]
    \centering
    \caption{An example of crafted sentence that causes DialoGPT to generate much longer outputs. 
    \textcolor{blue}{\textit{Italics}} and \textcolor{red}{\sout{strike through}} denote added and removed tokens, respectively.
    }
    \begin{tabular}{ll}
    \toprule
    \multicolumn{2}{l}{\begin{tabular}[c]{@{}l@{}}\textbf{Q}: Not really. Let's \textcolor{red}{\sout{talk}} \textcolor{blue}{\textit{think}} about food. What do you like to eat? \\I \textcolor{red}{\sout{love}} \textcolor{blue}{\textit{like}} fish. \end{tabular}} \\
    \multicolumn{2}{l}{\begin{tabular}[c]{@{}l@{}}\textbf{A}: I love fish too! What is your favorite kind? \end{tabular}} \\
    \multicolumn{2}{l}{\begin{tabular}[c]{@{}l@{}}$\hat{\mathbf{A}}$: I like to eat fish too. What is your favorite kind? I like pasta, filipino, \\steak, etc. I talk a lot on IRC and it is fun to learn about it with some \\other guys.\end{tabular}} \\
    \bottomrule
    \end{tabular}
    \label{tab:dialogue_adversary}
\end{table}

To evaluate the robustness of \Approach{} against malicious inputs, we apply a state-of-the-art adversarial attack method~\cite{li-etal-2023-white} that crafts provided input texts to elongate LM outputs. 
Table~\ref{tab:dialogue_adversary} presents an example of a malicious sentence designed to prompt an LM to generate longer output $\hat{\mathbf{A}}$ than the original one \textbf{A}, leading to a computational burst and degraded system performance.
Tasks are deemed malicious if their uncertainty scores exceed a  predefined threshold (see Eq.~\ref{eq:adv_threhsold}).
To assess the response, we control the proportion of deliberately crafted malicious tasks within a range of 0\% to 100\%, increasing in increments of 10\%, and evaluate subsequent system latency performance.

\begin{figure}
    \centering
    \includegraphics[width=0.45\textwidth]{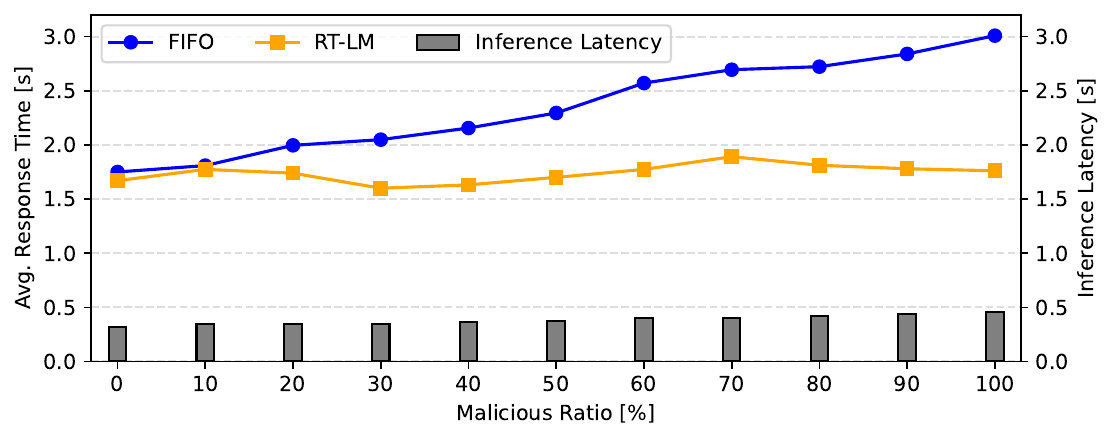}
    \caption{Average response time and LM inference latency on the edge server, under varying ratios of malicious tasks.}
    \label{fig:malicious}
    \vspace{-0.3cm}
\end{figure}

Fig.~\ref{fig:malicious} shows the effects of varying ratios of malicious tasks on the average response time of both FIFO and \Approach{}, as well as the associated average inference latency across different LMs.
As seen, \Approach{} is proficient in managing extreme conditions wherein a large proportion of malicious tasks need to be processed, outperforming the uncertainty-oblivious FIFO.
When the malicious task ratio exceeds 30\%, FIFO exhibits high sensitivity, with the average response time increases from around 2.0s to 3.0s. 
Whereas \Approach{} is resilient against malicious tasks, maintaining a steady average response time of around 1.5$\sim$1.9s.
Our results confirm that \Approach{} effectively prevents malicious tasks from hindering the execution of other critical tasks.
This resilience enhances \Approach{}'s suitability for applications like chatbots~\cite{zhou-etal-2020-design}, personal assistants~\cite{image-caption-generator}, and conversational AI in healthcare~\cite{bharti2020medbot} where defense against adversarial attacks is crucial.

\subsection{Overhead Analysis}
Analyzing overhead is crucial in practical real-time systems which are more complicated and variant.
A solution with high overhead may undermine response time and throughput, as the scheduling
process may severely block task execution. 
We present an analysis of both latency and memory usage introduced by \Approach{} on the edge server, offering insights into the practical efficiency of our design.

\noindent\textbf{Offline Profiling.}
We initialize an LW model and train it for 100 epochs, using the LM outputs as ground truths.
We report both the average training time per epoch and its proportion relative to the LM inference time.
Memory usage during this phase is also recorded.
As shown in Table~\ref{tab:offline}, our training consumes merely around 3$\sim$4\% of the LM inference latency, and less than 3\% of the total available memory (512 GB), demonstrating the overhead efficiency of \Approach{}.

\begin{table}[]
    \centering
    \caption{Latency and memory of offline profiling.
    }
    \resizebox{0.28\textwidth}{!}{%
    \begin{tabular}{c|cc|c}
    \toprule
        \multirow{2}{*}{LM} & \multicolumn{2}{c|}{Total LW latency (s)} & Memory \\
        \cmidrule{2-4}
        & Train & Ratio & Train \\
        \midrule
        DialoGPT & 351 & 3.01\% & 14,607 MB  \\
        GODEL & 490 & 3.96\% & 14,768 MB  \\
        BlenderBot & 448 & 3.71\% & 14,723 MB  \\
        BART & 392 & 3.25\% & 14,631 MB   \\
        T5 & 369 & 3.06\% & 14,639 MB  \\
    \bottomrule
    \end{tabular}}
    \label{tab:offline}
\end{table}

\begin{table}[]
    \centering
    \caption{Latency, memory, and CPU/GPU utilization of online scheduling.
    Prior., consol., and off. denote prioritization, consolidation, and offloading. 
    }
    \resizebox{0.48\textwidth}{!}{%
    \begin{tabular}{c|cccc|c|c}
    \toprule
        \multirow{2}{*}{LM} & \multicolumn{4}{c|}{Avg. per-task latency (ms)} & Memory & CPU / GPU util. \\
        \cmidrule{2-7}
         & Prior. & Consol. & Off. & Ratio & Test & Ratio \\
        \midrule
        DialoGPT & 8.04 & 0.42 & 0.37 & 2.10\% & 11,293 MB & 97\% / 92\% \\
        GODEL & 7.78 & 0.43 & 0.49 & 2.04\% &12,795 MB & 93\% / 97\% \\
        BlenderBot & 9.24 & 0.53 & 0.40 & 2.39\% & 12,136 MB & 99\% / 95\% \\
        BART & 7.84 & 0.35 & 0.10 & 2.06\% & 11,979 MB & 97\% / 91\% \\
        T5 & 8.39 & 0.33 & 0.18 & 2.27\% & 11,653 MB & 95\% / 90\%
        \\
    \bottomrule
    \end{tabular}
    }
    \label{tab:overhead}
    \vspace{-0.3cm}
\end{table}

\noindent\textbf{Online Scheduling.}
We evaluate the average per-task latency of each component of \Approach{} and compare the combined latency to the LM inference time.
We also record the average memory usage as well as CPU/GPU utilization during online scheduling.
Table~\ref{tab:overhead} reveals that \Approach{} introduces less than 3\% additional latency overhead relative to the LM inference time (around 415 milliseconds per task). 
Such small overheads are unlikely to affect real-time dialogue systems noticeably. 
Notably, prioritization accounts for the majority of scheduling time, as uncertainty is computed and queued at this stage.
For all LMs, CPU/GPU utilization reach over 90\%, which suggests effective resource allocation under \Approach{}.

\section{Related Work and Discussion}

\noindent\textbf{Real-time DNN Inference.} 
Recent research has improved real-time Deep Neural Network (DNN) performance with strategies optimizing performance-accuracy trade-offs~\cite{bateni2018apnet, zhou2018s, predjoule}, and exploring system design for DNN execution~\cite{neuos, DBLP:conf/rtss/KangLLSC21, xiang2019pipelined,DBLP:conf/rtss/NigadeBB022,jeong2022band,DBLP:conf/rtss/JiangLHHWXW21,DBLP:conf/rtss/JiYKASDK22,li2021efficient}. 
Despite these advancements, previous works neither consider the dynamics of DNNs for different execution times of inputs. In contrast, 
our proposed method, \Approach{}, builds upon these existing scheduling algorithms by incorporating uncertainty estimation to further enhance performance and resource allocation.

\noindent\textbf{Uncertainty Estimation.}
Uncertainty estimation has been a topic of interest in the machine learning and NLP community, particularly in the context of deep learning \cite{ovadia2019can}. 
Methods like Monte Carlo dropout \cite{gal-16-dropout} and Bayesian neural networks \cite{BNN} have been proposed to quantify the uncertainty in model predictions. 
Previous works~\cite{mao-etal-2021-eliciting,open-qa,li-etal-2023-uncertainty} also show that uncertainty may cause an LM to generate outputs with varied lengths. 
Our method employs a lightweight regressor to estimate the uncertainty in terms of the output length of an LM inference, which can be used to inform the scheduling process, improving resource utilization and response time.

\noindent\textbf{Intelligent Edge Server Systems.} In cloud-edge-client hierarchical systems, AI models are co-deployed on the cloud and edge servers~\cite{edge_connected_AI, AIGC_edge}, where multiple requests from diverse users via edge devices can be processed concurrently by the DNNs. Notable examples of such applications include online chatbots and live translation services. Additionally, cloud servers frequently grapple with load balancing across multiple workers~\cite{multi_workers}. \Approach{} could prioritize critical requests and redirect malicious tasks to CPU cores, thereby enhancing overall system performance and reducing the threat of performance attacks against DNNs~\cite{chen-etal-2023-dynamic,li-etal-2023-white,chen2022nicgslowdown,DBLP:journals/corr/abs-2306-00794,chen2022nmtsloth,chen2022deepperform}.

\noindent\textbf{Limitations of \Approach{}.}
\Approach{} mainly targets system-level optimization in heavy-workload scenarios, emphasizing concurrent task processing by taking into account the uncertainty characteristics of each task.
In real-world on-device LM-embedded systems, where queries typically arrive sequentially, there's room for further improvement, e.g., optimizing performance for each individual task by leveraging the correlation between uncertainty and layer-level LM inference/training efficiency could be pursued.
Additionally, our current approach is designed for single-machine scenarios. Expanding to hybrid deployment setups, such as server-edge combinations, is an avenue worth exploring.
Moreover, \Approach{} doesn't account for memory and power constraints, which could cause potential out-of-memory (OOM) issues on edge environments and pose challenges when deploying on low-power devices. Although deep learning compilers~\cite{chen2018tvm,chen2023dycl} may mitigate the challenges posed by limited resources in such scenarios, adapting \Approach{} to work efficiently in memory-constrained edge settings and optimizing LM inference from a power-efficiency standpoint is an area yet to be addressed.

\section{Conclusion}

In this paper, we introduced \Approach{}, a novel uncertainty-aware resource management for real-time on-device LMs. 
Our extensive evaluations demonstrated the superior performance of \Approach{} in terms of response time, system throughput, and robustness to various system settings, while maintaining low overhead and excellent memory efficiency. 
In the future, we will focus on further optimizing the uncertainty estimation mechanism and expanding the applicability of \Approach{} to more diverse and dynamic real-world workloads.

\section*{Acknowledgment}
This research was supported by the National Science Foundation under Grants CNS Career 2230968, CPS 2230969, CNS 2300525, CNS 2343653, CNS 2312397.

\bibliographystyle{IEEEtran}
\bibliography{IEEEabrv,ref}

\end{document}